# Evaluating the Robustness of Neural Language Models to Input Perturbations

**Milad Moradi, Matthias Samwald**
Institute for Artificial Intelligence
Medical University of Vienna, Austria
{milad.moradivastegani, matthias.samwald}@meduniwien.ac.at

## Abstract

High-performance neural language models have obtained state-of-the-art results on a wide range of Natural Language Processing (NLP) tasks. However, results for common benchmark datasets often do not reflect model reliability and robustness when applied to noisy, real-world data. In this study, we design and implement various types of character-level and word-level perturbation methods to simulate realistic scenarios in which input texts may be slightly noisy or different from the data distribution on which NLP systems were trained. Conducting comprehensive experiments on different NLP tasks, we investigate the ability of high-performance language models such as BERT, XLNet, RoBERTa, and ELMo in handling different types of input perturbations. The results suggest that language models are sensitive to input perturbations and their performance can decrease even when small changes are introduced. We highlight that models need to be further improved and that current benchmarks are not reflecting model robustness well. We argue that evaluations on perturbed inputs should routinely complement widely-used benchmarks in order to yield a more realistic understanding of NLP systems' robustness.

## 1 Introduction

High-performance deep neural language models such as BERT (Devlin et al., 2018), XLNet (Z. Yang et al., 2019), and GPT-2 (Radford et al., 2019) have brought breakthroughs to a wide range of Natural Language Processing (NLP) tasks including text classification, sentiment analysis, textual entailment, natural language inference, machine translation, and question answering. Their immense ability in capturing various linguistic properties has led these state-of-the-art language models to master different NLP tasks, even surpassing human accuracy on some benchmarks such as SQuAD[1].

However, recent studies have revealed that there is a gap between performing well on benchmarks and actually working under real-world situations (Belinkov and Bisk, 2018; Ribeiro et al., 2020). Even a well-trained, high-performance deep language model can be sensitive to negligible changes in the input that cause the model to make erroneous decisions (M. Sun et al., 2018). This raises serious concerns regarding the robustness/reliability of neural language models utilized in real-world applications. The terms 'robustness' and 'reliability' refer to the ability of a system to perform consistently well in situations where changes to input should not cause a change in the system's output, or the system is expected to properly reflect the change and produce a correct outcome.

Applying automatic or human-controlled perturbations to textual inputs has been shown to be effective for evaluating the robustness of NLP systems, investigating their vulnerabilities, and finding their bugs. Recently, CheckList (Ribeiro et al., 2020) provided a framework for behavioral testing of NLP systems inspired by black-box testing in software engineering. CheckList enabled generating new (perturbed) test samples through abstracting different test types aimed at testing linguistic capabilities. Other studies focused on evaluating robustness to perturbed inputs for machine translation (Belinkov and Bisk, 2018; Niu et al., 2020), perturbation sensitivity analysis for detecting unintended model biases (Prabhakaran et al., 2019), or robustness to adversarial perturbations (Alshemali and Kalita, 2020; Ebrahimi et al., 2018; Liang et al., 2018). However, a comprehensive methodology for evaluating the performance of NLP models under real-world conditions is still missing.

---

[1] https://rajpurkar.github.io/SQuAD-explorer/

In a realistic scenario, the input text may contain typos and misspellings that should not cause any changes in the NLP system's outcome. Minor grammatical errors may appear in the text, but the semantics is still preserved, therefore, the NLP system is expected to treat the input as it was error-free. Some deliberate or unintentional changes may modify the semantics, and the NLP model is expected to reflect the changes in the outcome. These are only few examples of natural noise in text data that NLP systems should have the ability to properly deal with.

In this paper, we design and implement a wide range of character-level and word-level systematic perturbations to textual inputs in order to simulate different types of noise that a NLP system may face in real-world use cases. Conducting extensive experiments on various NLP tasks, we investigated the ability of four neural language models, i.e. BERT, RoBERTa, XLNet, and ELMo, in handling slightly perturbed inputs. The results reveal that the neural models are unstable to small changes that can be easily handled by humans, e.g. misspellings, missing words, repeated words, synonyms, etc. The systematic input perturbations can expose the vulnerabilities of NLP systems and bring more insights into how high-performance models behave when they encounter noisy yet understandable inputs. This study suggests that the performance of NLP models should not be overestimated by only relying on accuracy scores obtained on benchmark datasets.

Similar to CheckList, our perturbation framework treats NLP systems as black-boxes. This facilitates comparison of different models, without needing to know the model structure and internals. CheckList focuses on testing linguistic capabilities of NLP systems, e.g. handling coreferences, identifying named entities, semantic role labeling, and vocabulary. On the other hand, our perturbation methods aim at evaluating the robustness of NLP systems to noisy inputs. Our input perturbation framework can act as a complement to the CheckList testing methodology.

In CheckList, many test types rely on creating synthetic samples from scratch by the user, which is a time-consuming task, and needs much creativity and effort. Moreover, synthetic samples may suffer from low coverage (Ribeiro et al., 2020). However, most of the perturbation methods introduced in this paper do not need human intervention; they can automatically generate perturbed samples that still preserve the semantics and are sufficiently meaningful to users.

Some types of perturbation utilized in this work were already tested in previous work on adversarial attacks on NLP systems (Zeng et al., 2020; Zhang et al., 2020). However, adversarial perturbations are considered worst-case scenarios that do not occur frequently in real-world situations, representing a very specific type of noise (Fawzi et al., 2016). In order to generate effective adversarial examples, most attack methods need to have access to the NLP model structure, internal weights, and hyperparameters, which may not be possible in every testing scenario (Zhang et al., 2020). Furthermore, adversarial perturbations should not be perceived by humans (Liang et al., 2018). This is a serious challenge, since even small changes to a text may be easily recognized by the user.

To the best of our knowledge, this paper is the first study that presents empirical results achieved with a comprehensive set of non-adversarial perturbation methods for testing robustness of NLP systems on non-synthetic text. An important contribution of this work is to evaluate the robustness of several high-performance language models on various NLP tasks using different types of character-level and word-level input perturbations. Moreover, to ascertain the usefulness of the perturbations (i.e. how effectively they can be used to automatically generate meaningful and understandable perturbed samples), we conducted an extensive user study.

## 2 NLP tasks

In our experiments, we used five datasets covering five different NLP tasks. Table 1 summarizes some statistics of the datasets. A short description of the datasets is given in the following.

**TREC** (Li and Roth, 2002) is a Text Classification (TC) dataset containing more than 6,000 questions and 50 different class labels that specify the type of questions.

**Stanford Sentiment Treebank (SST)** (Socher et al., 2013) is a Sentiment Analysis (SA) dataset containing more than 11,000 movie reviews from 'Rotten Tomatoes'. Every review is classified into one of the five classes: *very positive*, *positive*, *neutral*, *negative*, and *very negative*.

**CoNLL-2003**[2] is a Named Entity Recognition (NER) dataset containing news stories from the Reuters corpus with more than 200K tokens annotated as *Person*, *Organization*, *Location*, *Miscellaneous*, or *Other*.

**STS benchmark** (Cer et al., 2017) is a Semantic Similarity (SS) dataset comprising of more than 8K text pairs extracted from image captions, news headlines, and user forums. Each pair of sentences is assigned a similarity score between 0 and 5.

**WikiQA (WQA)** (Y. Yang et al., 2015) is a Question Answering (QA) dataset composed of more than 3,000 questions and 29,000 sentences as answers extracted from Wikipedia.

## 3 Language models

In our experiments, we utilized four neural language models shown to be effective in learning bidirectional contexts and obtained state-of-the-art results during recent years:

**BERT** (Devlin et al., 2018) is composed of deep encoder transformer layers and uses two pretraining objectives, i.e. masked language modelling and next sentence prediction. We used the BERT$_{LARGE}$ architecture (along with the *cased* model) containing 24 transformer layers, 1024 hidden units per layer, 16 attention heads per hidden unit, and 340 million parameters.

**RoBERTa** (Liu et al., 2019) uses a model architecture similar to BERT, but adopts an optimized pretraining approach. It was pretrained on more data, with bigger batch sizes and longer sequences than BERT. Furthermore, the next sentence prediction objective was removed and a dynamic masking strategy replaced the basic masking method. We used RoBERTa$_{LARGE}$ that further optimizes the same model as BERT$_{LARGE}$.

**XLNet** (Z. Yang et al., 2019) utilizes decoder transformers and adopts a permutation language modelling approach along with generalized autoregressive pretraining. We used the XLNet$_{LARGE}$ model, with the same architecture hyperparameters and model size as BERT$_{LARGE}$.

**ELMo** (Peters et al., 2018) is a contextualized word representation method that utilizes character convolutions along with shallow concatenation of backward and forward LSTMs to implement bidirectional language modeling. We used the original ELMo model composed of two highway layers with an LSTM hidden size of 4096, output size of 512, and a total parameters of 93.6 million. The contextualized embeddings computed by ELMo were fed into a dense layer containing 128 hidden units followed by an output layer with a softmax activation in the TC and QA tasks, a linear activation in the SA and SS tasks, and CRF layer with a linear activation in the NER task.

| Dataset | Task | Train | Dev | Test | Eval. Measure |
|---|---|---|---|---|---|
| TREC | TC | 5,000 | 452 | 500 | Micro F1-score |
| SST | SA | 8,544 | 1,101 | 2,210 | Accuracy |
| CoNLL | NER | 14,041 | 3,250 | 3,453 | F1-score |
| STS | SS | 5,749 | 1,500 | 1,379 | Pearson |
| WQA | QA | 2,117 | 296 | 630 | F1-score |

Table 1: The main statistics of the datasets used to conduct the perturbation experiments.

We retrieved the pretrained models, fine-tuned them separately on each downstream task using the training and development sets, and tested them on the test sets. We utilized the Huggingface transformers (Wolf et al., 2020) and FARM [3] libraries to implement the transformer-based models. A complete list of hyperparameter values is presented in Appendix A.

## 4 Perturbation methods

We designed and implemented various character-level and word-level perturbation methods that simulate different types of noise an NLP system may encounter in real-world situations. The perturbations can be produced for every dataset regardless of the underlying language model or NLP system being tested. Table 2 presents an example for every perturbation method. The perturbation methods were implemented in Python using the NLTK library. The source code is available at https://github.com/mmoradi-iut/NLP-perturbation.

---

[2] https://github.com/synalp/NER/tree/master/corpus/CoNLL-2003

[3] https://github.com/deepset-ai/FARM

| Perturbation | Original text | Perturbed text |
|---|---|---|
| *Character-level* | | |
| Insertion | Who was the first governor of Alaska? | Who was the firsdt governor of Alaska? |
| Deletion | Mercury, what year was it discovered? | Mercury, what year was it discovred? |
| Replacement | Who is the Prime Minister of Canada? | Who is the Prime Monister of Canada? |
| Swapping | What is the primary language in Iceland? | What is the primary lnaguage in Iceland? |
| Repetition | How many hearts does an octopus have? | How many heartts does an octopus have? |
| CMW | What kind of gas is in a fluorescent bulb? | What kind of gas is in a florescent bulb? |
| LCC | How many hearts does an octopus have? | How many hearts does an OCTOPUS have? |
| *Word-level* | | |
| Deletion | How much was a ticket for the Titanic? | How much a ticket for the Titanic? |
| Repetition | What is another name for vitamin B1? | What is another name name for vitamin B1? |
| RWS | What precious stone is a form of pure carbon? | What valued rock is a form of pure carbon? |
| Negation | What planet is known as the "red" planet? | What planet is not known as the "red" planet? |
| SPV | What does a barometer measure? | What do a barometer measure? |
| Verb tense | Why in tennis are zero points called love? | Why in tennis were zero points called love? |
| Word order | What is the most common eye color? | What is the common most color eye? |

Table 2: Character-level and word-level perturbation examples from the TREC question classification dataset. CMW: Common Misspelled Words, LCC: Letter Case Changing, RWS: Replacement With Synonyms, SPV: Singular/Plural Verbs.

Almost all the character-level perturbations presented here were already tested in adversarial attack scenarios (Heigold et al., 2018; Zeng et al., 2020; Zhang et al., 2020), but were not yet implemented in a non-adversarial testing framework, except the misspelling perturbation implemented by CheckList. Among the word-level perturbations, *Deletion*, *Repetition*, *Singular/plural verbs*, *Word order*, and *Verb tense* were not already used to test the robustness. However, *Negation* was included in CheckList, and *Replacement with Synonyms* was used for adversarial attack (Dong et al., 2020; Ren et al., 2019).

### 4.1 Character-level perturbation

These perturbation methods randomly select a word, denoted as $Word_i$, and apply perturbations to its characters. They are described in the following.

**Insertion**. A character is randomly selected and inserted in a random position (except the first and last position) if $Word_i$ contains at least three characters.

**Deletion**. A character is randomly selected and deleted if $Word_i$ contains at least three characters. The last and first characters of $Word_i$ are never deleted.

**Replacement**. A character is randomly selected and is replaced by an adjacent character on the keyboard.

**Swapping**. A character is randomly selected and swapped with the adjacent right or left character in $Word_i$.

**Repetition**. A character in a random position (except the first and last position) is selected and a copy of it is inserted right after the selected character.

**Common misspelled words**. If a word in the input text appears in the Wikipedia corpus of common misspelled words[4], it is replaced by its misspelling.

**Letter case changing** toggles the letter case, i.e. converts a lower case character to its upper case form and vice versa. The letter case changing is done for either the first or all the characters of $Word_i$. The type of letter case changing is specified in a random manner.

### 4.2 Word-level perturbation

**Deletion** randomly selects a word from the input sample and removes it.

**Repetition** selects a random word, makes a copy of it, and inserts it right after the selected word.

**Replacement with synonyms** replaces words contained in the sample by their synonyms extracted from the WordNet lexical database (Miller, 1995).

**Negation**. It identifies verbs in the sample, then injects negations by converting positive verbs to negative, or removes negation by converting negative verbs to positive. The goal is to investigate the ability of the NLP system in adapting its outcome to reflect the injected or removed negation.

---

[4] https://en.wikipedia.org/wiki/Wikipedia:Lists_of_common_misspellings/For_machines/

| Task | LM | Test set | Character-level perturbation methods | | | | | | |
|---|---|---|---|---|---|---|---|---|---|
| | | | Insertion | Deletion | Replace | Swap | Repeat | CMW | LCC |
| TC | BERT | 90.4 | 77.4 | 76.2 | 76.1 | 76.5 | 78.8 | 58.4 | 78.3 |
| | RoBERTa | **93.1** | 79.2 | **78.9** | 76.3 | **76.7** | 80.8 | 60.5 | 78.9 |
| | XLNet | 92.0 | 78.1 | 78.3 | **76.5** | 75.2 | 80.2 | 61.5 | 77.4 |
| | ELMo | 84.8 | **80.4** | 78.5 | 74.7 | 75.6 | 79.6 | **61.9** | **80.8** |
| SA | BERT | 92.2 | 77.1 | 75.6 | 75.5 | **78.3** | 77.9 | 62.0 | 76.7 |
| | RoBERTa | **94.0** | 79.3 | 76.8 | 75.1 | 76.2 | 79.3 | 64.0 | 78.3 |
| | XLNet | 93.1 | 78.3 | 78.7 | 75.9 | 73.8 | **81.1** | **65.6** | 78.9 |
| | ELMo | 87.6 | **79.7** | **79.0** | 76.2 | 78.1 | 78.4 | 64.1 | **79.1** |
| NER | BERT | 92.6 | 83.6 | 80.7 | 81.4 | 82.5 | 81.9 | 71.3 | 81.2 |
| | RoBERTa | **93.3** | **84.3** | 80.9 | **81.7** | **83.1** | **82.4** | **71.8** | **81.5** |
| | XLNet | 92.7 | 83.9 | **81.1** | 81.3 | 82.7 | 81.5 | 71.6 | 81.3 |
| | ELMo | 90.2 | 83.0 | 80.2 | 80.9 | 82.2 | 81.3 | 70.8 | 81.0 |
| SS | BERT | 82.5 | 72.9 | 71.5 | 73.0 | 74.3 | 74.2 | 68.6 | 73.8 |
| | RoBERTa | **83.9** | **73.5** | **72.8** | **73.6** | **75.1** | **74.7** | **69.5** | **74.9** |
| | XLNet | 83.3 | 73.3 | 72.0 | 73.2 | 74.6 | 74.1 | 67.9 | 74.4 |
| | ELMo | 80.7 | 71.1 | 70.9 | 72.3 | 73.8 | 72.5 | 67.0 | 72.6 |
| QA | BERT | 91.6 | 82.7 | 80.5 | 81.1 | 81.9 | 79.8 | 68.6 | 80.7 |
| | RoBERTa | **94.9** | **84.1** | **81.7** | **82.9** | **83.2** | 81.6 | **72.5** | **84.0** |
| | XLNet | 93.4 | 83.5 | 81.1 | 82.3 | 81.9 | **82.8** | 71.5 | 83.3 |
| | ELMo | 85.5 | 80.6 | 79.5 | 76.0 | 78.3 | 80.1 | 67.9 | 81.1 |

Table 3: Performance of the language models on the test sets and character-level perturbed samples of the downstream tasks. For every task and every perturbation method, the highest score is shown in bold face. CMW: Common Misspelled Words, LCC: Letter Case Changing.

This perturbation method operates based on a set of rules that assess verbs, subjects, and verb tenses based on POS tags, then applies an appropriate rule to construct the test sample. For example, if the POS tag of a verb is VBZ, the verb appears in the third person simple present form. Therefore, the verb is replace by [does not + VBP] where VBP is the basic form of the verb, in order to inject negation into the sample.

**Singular/plural verbs**. It simulates a common error in real use cases, i.e. using plural form of a verb instead of the singular form, and vice versa, usually with a third-person subject. This perturbation does not usually change the text's meaning in most NLP tasks if the task does not rely on the subject-verb agreement. Therefore the NLP system should treat the perturbed sample as an unperturbed text.

**Word order**. It randomly selects *M* consecutive words from the sample and changes the order in which they appear in the text. The goal is to investigate whether the NLP system is sensitive to word ordering or it only decides based on the presence of words in the input.

**Verb tense**. It converts present simple or continuous verbs to their corresponding past simple or continuous forms, or vice versa. The goal is to assess the sensitivity of the NLP system to changing the verb tense in tasks where the verb tense is not important to the output. In this case, the system's output should not change after modifying the verb tense. This method first extracts POS tags to identify verbs and their subjects. It then converts the verb tense using the `mlconjug3` package and reconstruct the sentence with the new verb tense.

## 5 Experimental results

All the experiments were performed on a computer with an Intel Core i5-9600K CPU at 3.70GHz, 32 GB of RAM, and a GeForce RTX 2080 Ti graphic card (GPU) with 11 GB dedicated memory. Perturbation methods ran on CPU; fine-tuning on training sets, and evaluating on test sets and perturbed samples ran on GPU.

### 5.1 Performance on perturbed inputs

Since it has been proven that sentences that contain few typos, misspellings, or minor character-level errors can be still fully understandable to humans (Belinkov and Bisk, 2018; Xu and Du, 2020), character-level perturbations are not expected to change the text's meaning in most cases. Therefore, they can be automatically produced and used for testing the robustness of NLP systems.

On the other hand, some word-level perturbations may change the text's meaning.

| Task | LM | Test set | Word-level perturbation methods | | | | | | |
|---|---|---|---|---|---|---|---|---|---|
| | | | Deletion | Repeat | RWS | Negation | SPV | VT | WO |
| TC | BERT | 90.4 | 75.1 | **89.3** | 65.7 | 89.1 | 88.2 | 89.0 | 74.5 |
| | RoBERTa | **93.1** | **76.2** | 88.7 | 73.2 | **90.3** | **89.5** | 89.4 | 78.5 |
| | XLNet | 92.0 | **76.2** | 87.5 | 72.7 | 89.4 | 89.0 | **89.6** | **83.1** |
| | ELMo | 84.8 | 72.9 | 82.8 | **75.1** | 83.5 | 83.6 | 81.2 | 62.9 |
| SA | BERT | 92.2 | 73.7 | 87.6 | 67.5 | **84.6** | 88.2 | **90.1** | 76.4 |
| | RoBERTa | **94.0** | 74.5 | **90.1** | 74.2 | 83.9 | **88.7** | 88.6 | 77.5 |
| | XLNet | 93.1 | **74.7** | 88.5 | 74.1 | 82.3 | 88.6 | 89.3 | **83.8** |
| | ELMo | 87.6 | 72.0 | 80.6 | 73.1 | 75.4 | 84.6 | 82.9 | 65.9 |
| NER | BERT | 92.6 | 81.4 | 83.1 | 74.1 | 85.3 | 88.2 | 89.1 | 70.7 |
| | RoBERTa | **93.3** | **82.3** | **83.9** | **74.5** | **85.8** | **88.6** | **89.4** | 71.1 |
| | XLNet | 92.7 | 81.9 | 83.7 | 73.9 | 85.6 | 88.3 | 88.7 | **74.8** |
| | ELMo | 90.2 | 79.7 | 82.1 | 69.3 | 82.4 | 85.1 | 84.9 | 68.5 |
| SS | BERT | 82.5 | 72.6 | 74.1 | 69.4 | 68.5 | 75.2 | 75.6 | 72.0 |
| | RoBERTa | **83.9** | **74.1** | **74.8** | 70.0 | **69.2** | 75.7 | **76.7** | 73.9 |
| | XLNet | 83.3 | 73.3 | 74.5 | 69.8 | 68.7 | **75.8** | 76.2 | **75.3** |
| | ELMo | 80.7 | 69.8 | 71.7 | 67.4 | 66.0 | 73.2 | 72.8 | 72.6 |
| QA | BERT | 91.6 | 78.4 | **89.6** | 71.5 | 84.9 | 88.0 | 90.3 | 76.7 |
| | RoBERTa | **94.9** | **79.9** | 89.3 | **78.1** | **86.5** | **89.7** | **91.2** | 79.0 |
| | XLNet | 93.4 | 79.2 | 89.5 | 77.3 | 86.1 | 89.1 | 90.9 | **85.8** |
| | ELMo | 85.5 | 73.5 | 81.4 | 75.0 | 82.7 | 84.1 | 81.9 | 67.3 |

Table 4: Performance of the language models on the test sets and word-level perturbed samples of the downstream tasks. For every task and every perturbation method, the highest score is shown in bold face. For three perturbation methods, i.e. *Deletion*, *Negation*, and *RWS*, 200 perturbed samples were used in the experiments. RWS: Replacement With Synonyms, SPV: Singular/Plural Verbs, VT: Verb Tense, WO: Word Order.

Consequently, the perturbed samples should be monitored to make sure they are still meaningful with respect to the NLP task at hand, and are consistent with the original label in the dataset. Otherwise, they should not be used for testing the robustness, or the label should be changed to reflect the change and preserve the consistency.

We separately applied every character-level perturbation method to all test samples in a dataset, and all the resulting perturbed samples were used to evaluate the robustness of the language models. A hyperparameter named Perturbation Per Sample (PPS) specified the maximum number of perturbations in a sample.

We monitored and filtered perturbed samples resulted from three word-level perturbations that may change the text's meaning. These perturbations are *Deletion*, *Negation*, and *Replacement with synonym*. For every sample whose meaning was changed by these three methods, and a change in the test set label was necessary to preserve consistency, we altered the label if it was applicable. If a proper label could not be assigned to the perturbed sample or the resulting text was no longer meaningful, we excluded the sample from the evaluations. Since monitoring and filtering every single perturbed sample was extremely time-consuming (such that approximately one minute was needed on average to check the meaningfulness of a perturbed sample and its consistency with the test set label), we corrected labels and filtered perturbed samples for the above three methods until 200 samples were collected for every dataset; then we used these samples to evaluate the models on perturbed inputs. We performed this manual curation of perturbed samples for all values of PPS that we experimented, i.e. values in the range [1, 4]. Appendix B presents the number of perturbed samples checked in the manual curation procedure until reaching 200 test samples for every dataset and different values of PPS. The manual curation was performed by three annotators who had sufficient English language knowledge to properly judge about the meaningfulness and consistency of perturbed samples.

Since the rest of word-level perturbations are not expected to change the text's meaning with respect to the NLP tasks in our experiments, we did not monitor and filter them; they were produced and used automatically. Again, the PPS hyperparameter controlled the maximum number of perturbations in every sample.

Table 3 and Table 4 present the performance of the language models on character-level and word-level perturbed samples, respectively. These results are reported for PPS=1. The performance of the language models on original, unperturbed test sets is also reported in both tables for every NLP task. We performed five separate fine-tuning runs to test if the performance of the NLP models on the original test set and perturbed samples vary between individual runs. Since there was no statistically significant difference between multiple runs (with respect to a t-test with a significance level of $p$=0.05), we only report the results of the first fine-tuning and testing run. The language models were neither pretrained nor fine-tuned on perturbed samples. The perturbation methods were only applied to the test sets. As the results show, the language models are sensitive to the perturbations and their performance decreases when the input is slightly noisy. However, RoBERTa still performs better than the other models, and ELMo obtains the lowest scores in general.

|      |         | Character-level perturbations |       |       |       | Word-level perturbations |       |       |       |
|------|---------|:-----:|:-----:|:-----:|:-----:|:-----:|:-----:|:-----:|:-----:|
| Task | LM      | PPS=1 | PPS=2 | PPS=3 | PPS=4 | PPS=1 | PPS=2 | PPS=3 | PPS=4 |
| TC   | BERT    | −15.8 | −17.2 | −18.0 | −18.3 | −8.8  | −10.2 | −13.1 | −13.8 |
|      | RoBERTa | −17.1 | −17.9 | −18.5 | −18.9 | −9.4  | −11.0 | −12.7 | −13.3 |
|      | XLNet   | −16.6 | −17.4 | −19.2 | −19.7 | −8.0  | −10.3 | −12.4 | **−12.9** |
|      | ELMo    | **−8.8** | **−10.0** | **−11.2** | **−11.8** | **−7.3** | **−9.1** | **−11.5** | −13.2 |
| SA   | BERT    | −17.4 | −18.9 | −20.1 | −21.3 | −11.0 | −13.5 | −15.1 | −16.7 |
|      | RoBERTa | −18.4 | −19.7 | −21.2 | −21.9 | −11.5 | −12.9 | **−14.2** | **−14.8** |
|      | XLNet   | −17.0 | −19.4 | −20.6 | −21.7 | **−10.0** | **−12.4** | −14.3 | −15.6 |
|      | ELMo    | **−11.2** | **−14.4** | **−16.0** | **−17.5** | −11.2 | −13.8 | −14.9 | −16.1 |
| NER  | BERT    | −12.2 | −14.6 | −16.3 | −16.9 | −10.8 | −12.7 | −14.0 | −14.9 |
|      | RoBERTa | −12.4 | −14.0 | −14.8 | −16.7 | −11.0 | −12.5 | −13.6 | −14.3 |
|      | XLNet   | −12.2 | −13.8 | −14.5 | −15.0 | **−10.2** | **−11.9** | **−13.1** | **−14.0** |
|      | ELMo    | **−10.2** | **−12.5** | **−13.1** | **−13.8** | −11.3 | −13.0 | −15.2 | −16.1 |
| SS   | BERT    | −9.8  | −11.3 | −12.9 | −13.6 | −10.0 | −12.2 | −13.8 | −14.5 |
|      | RoBERTa | −10.4 | −11.8 | −13.0 | −14.3 | −10.4 | −12.1 | −13.2 | −14.1 |
|      | XLNet   | −10.5 | −12.0 | −13.2 | −14.1 | **−9.9** | **−11.4** | **−12.8** | **−13.3** |
|      | ELMo    | **−9.2** | **−10.6** | **−11.7** | **−13.2** | −10.2 | −12.3 | −13.9 | −15.4 |
| QA   | BERT    | −12.2 | −14.2 | −15.0 | −16.4 | −8.8  | −10.2 | −12.5 | **−13.0** |
|      | RoBERTa | −13.4 | −15.1 | −15.8 | −16.5 | −10.1 | −11.0 | −12.3 | −13.5 |
|      | XLNet   | −12.4 | −13.5 | −16.4 | −18.9 | −7.9  | **−9.5** | **−11.2** | −13.1 |
|      | ELMo    | **−7.8** | **−9.7** | **−11.3** | **−12.0** | **−7.5** | −10.1 | −12.0 | −13.9 |

Table 5: Absolute decrease in the performance of the language models, on different NLP tasks, for character-level and word-level perturbations, and with different values of the hyperparameter Perturbation per Sample (PPS). For every task and every value of the hyperparameter PPS, the lowest decrease in the performance is shown in bold face.

The results suggest that some language models can handle specific types of perturbation more effectively than other models. ELMo obtains higher scores than BERT and even performs on par with XLNet and RoBERTa on some character-level perturbations. This can be due to its pure character-based representation that enables the model to use morphological clues, leading to a more robust model against character-level noises.

XLNet is shown to handle perturbations to word ordering more efficiently than the others. This can be an effect of the permutation language modelling that may allow the model to still capture the context and perform more accurately when some context words appear in a different order. The results also suggest those models that were pretrained on larger corpora such as RoBERTa and XLNet are more robust when words are replaced by their synonyms. Furthermore, when the negation perturbation has more impact on the task at hand, e.g. sentiment analysis, the models are less stable and handle the noise less efficiently than on other tasks. Observing the results, we can also point out the LSTM-based model, i.e. ELMo, is more sensitive to the order of words in a sample than the transformer-based models.

Table 5 presents the absolute decrease in the performance of the language models for different PPS values in the range [1, 4]. For every language model, the average of absolute decrease in performance is separately reported on character-level and worl-level perturbations for every NLP task. As can be shown, the models are generally more sensitive to character-level perturbations than word-level ones. Perturbed inputs causes the models to make erroneous outcomes on the sentiment analysis task more often than on the other tasks. On the other hand, the question answering task suffers less than the other tasks from noisy inputs.

Figure 1 represents six examples for which perturbations to the input led the RoBERTa model

**Example 1) Char-level: Misspelling**
Text classification: TREC

**Input:** What is the lenght of the caostline of the state of Alaska ?
**Label:** *NUMBER:distance*
**Predicted:** *LOCATION:state*

**Example 2) Char-level: Misspelling**
Semantic similarity: STS benchmark

**Input:** A small bird sitting on a branch in witner.
A small bird perched on an icy brnach.
**Similarity:** *4.2*
**Estimated:** *3.5*

**Example 3) Char-level: Swapping**
Question answering: WikiQA

**Question:** What deos alkali do to liqiuds?
**Answer:** *Some authors also define an alkali as a base that dissolves in water.*
**Predicted:** *A solution of a soluble base has a pH greater than 7.*

**Example 4) Word-level: Replace with synonym**
Text classification: TREC

**Input:** What is the diam of a golf_game ball ?
**Label:** *NUMBER:distance*
**Predicted:** *DESCRIPTION:definition*

**Example 5) Word-level: Repetition**
Semantic similarity: STS benchmark

**Input:** A man plays a trumpet trumpet.
A man is playing playing the trumpet.
**Similarity:** *5.0*
**Estimated:** *4.3*

**Example 6) Word-level: Deletion**
Question answering: WikiQA

**Question:** what is name of national (anthem) song of Switzerland?
**Answer:** *The Swiss Psalm is the national anthem of Switzerland.*
**Predicted:** *Since then, it has been frequently sung at patriotic events.*

Figure 1: Six examples of input perturbations from the three NLP tasks for which the RoBERTa model made wrong decisions, but it made correct decisions on the respective original inputs.

to make wrong decisions, but the model made correct decisions on the respective original inputs. As can be seen, examples 1-3 contain minor character-level noise that causes the model make wrong decisions, however, the perturbed text still seems understandable. In example 4, 'diameter' was replaced by 'diam' and 'golf' was replaced by 'golf_game', but the model failed to handle these changes. In example 5, two repetitive words led the model to estimate a lower similarity score, however, the semantic remained unchanged. Finally, example 6 shows how removing a single word led the model to choose a wrong answer.

## 5.2 User study

We conducted a user study with 20 participants to investigate how understandable the perturbed texts are to humans. We created a set of perturbations by randomly selecting perturbed samples from the datasets used in the experiments. The samples covered all types of character-level and word-level perturbations.

In the first part of the study, each participant was given 30 perturbed samples from those perturbation methods that are not expected to change the text's meaning with respect to the NLP tasks at hand. These are all the character-level perturbations and three word-level perturbations, i.e. *Repetition*, *Singular/plural verbs*, and *Verb tense*. The participants were also given the original text along with every perturbed sample, and were asked to judge if the perturbed text is understandable and still conveys the same meaning. Every sample contained one, two, or three perturbations.

According to the user evaluations, on average, 94% of the perturbed samples from this set were understandable and still conveyed the same meaning as the original text. These results are well in agreement with our discussion in Section 6.2, i.e. the majority of our proposed perturbations can be automatically produced and used without needing human supervision to ensure understandability and consistency.

In the second part of the study, each participant was given 20 perturbed samples from those perturbation methods that may change the text's meaning or result in meaningless text. They are the rest of word-level perturbations, i.e. *Deletion*, *Replacement with synonyms*, *Negation*, and *Word order*. The participants were also given the original text along with every perturbed sample, and were asked to judge (with respect to the task at hand) if the perturbed text is still meaningful and consistent with the test set label.

According to the user evaluations, on average, 39% of the perturbed samples from this set were still meaningful and consistent with the label, 12% of the perturbed samples were meaningful but the label should be changed, and 49% of the perturbed samples were no longer meaningful. These results imply that some perturbations need to be monitored, corrected, or filtered to make sure they are understandable, meaningful, and consistent with the test set label. This helps to fairly estimate the robustness of NLP systems to input perturbations.

## 6 Related work

Typical performance measures such as accuracy, precision, recall, etc. may not properly reflect how NLP systems behave in real-world use cases. This has motivated many studies to devise novel methods for investigating different capabilities and vulnerabilities of text processing systems. Behavioral testing introduces targeted changes to textual inputs to test linguistic capabilities of systems (Ribeiro et al., 2020). Explanations provide simplified representations of what a complex NLP model has learned (Moradi and Samwald, 2021a, b; Ribeiro et al., 2016). This can help to identify biases and errors in NLP models. Adversarial perturbations have been widely studied to assess the robustness of NLP systems against adversarial samples crafted to fool a model (Alshemali and Kalita, 2020; Ren et al., 2019; Zhang et al., 2020). However, adversarial samples resemble a very specific type of noise. Moreover, most of previous work on adversarial perturbation to NLP models focused on misspelling attacks (Jones et al., 2020; Pruthi et al., 2019; L. Sun et al., 2020). The perturbation methods implemented in this paper represented a wide range of noises that an NLP system may face in real-world situations.

Introducing noise and changing textual inputs were already adopted to assess the ability of models in capturing specific linguistic features such as learning syntax-sensitive dependencies (Linzen et al., 2016), for specific NLP tasks such as machine translation (Belinkov and Bisk, 2018), for detecting biases in language models (Prabhakaran et al., 2019), or to identify susceptible entities in text documents (M. Sun et al., 2018). In this paper, we investigated the robustness on a wide range of tasks, and for various types of character-level and word-level noises in text.

## 7 Conclusion

In this paper, we introduced and implemented a set of non-adversarial perturbation methods that can be used to evaluate the robustness of NLP systems. We extensively investigated the robustness of high-performance neural language models to noisy input texts. The evaluations on various NLP tasks imply that these models are sensitive to different character-level and word-level perturbations to the input, and the models' performance can decrease when the input contains slight noise. The results suggest that it may be too simplistic to only rely on accuracy scores obtained on benchmark datasets when evaluating the robustness of NLP systems.

The proposed perturbations can be used, along with other methodologies such as CheckList, to test how robust and reliable NLP systems can operate in real-world settings. The experimental results demonstrated that the perturbation methods are effective tools for evaluating NLP systems against noisy data. The user study revealed that only few perturbation methods need to be monitored to make sure they produce meaningful and consistent samples. Most of the perturbation methods can be used automatically to produce noisy test samples. They can be also used as a baseline for evaluating adversarial attacks against non-adversarial perturbations.

Future work may include helping users assess meaning preservation and grammatical correctness in a semi-automatic manner. Sentence encoders such as InferSent (Conneau et al., 2017), Universal Sentence Encoder (Cer et al., 2018), and BERT trained for semantic similarity (Reimers and Gurevych, 2019) can be used to give users clues how semantically similar the original and perturbed sentences are. Moreover, users can be provided with information about grammatical errors in the perturbed text using LanguageTool (Naber, 2003) or other grammar checking tools.

**Appendix A. Fine-tuning hyperparameters**

Table 6, 7, 8, and 9 present the hyperparameter values of the BERT, RoBERTa, XLNet, and ELMo models, respectively, in the fine-tuning experiments on different NLP tasks. Those hyperparameters not included in the tables were used with their default values specified by the original models.

| Hyperparameter | TC | SA | QA | NER | SS |
|---|---|---|---|---|---|
| Max sequence length | 64 | 64 | 256 | 64 | 64 |
| Batch size | 12 | 12 | 2 | 12 | 12 |
| Learning rate | 2e-5 | 1e-5 | 3e-5 | 1e-5 | 2e-5 |
| Num epochs | 20 | 15 | 10 | 20 | 20 |

Table 6: Fine-tuning hyperparameters for the BERT language model on different tasks. TC: Text Classification, SA: Sentiment Analysis, QA: Question Answering, NER: Named Entity Recognition, SS: Semantic Similarity.

| Hyperparameter | TC | SA | QA | NER | SS |
|---|---|---|---|---|---|
| Max sequence length | 64 | 64 | 256 | 64 | 64 |
| Batch size | 12 | 12 | 2 | 12 | 12 |
| Learning rate | 1e-5 | 1e-5 | 1.5e-5 | 2e-5 | 2e-5 |
| Weight decay | 0.1 | 0.1 | 0.01 | 0.1 | 0.01 |
| Learning rate decay | Linear | Linear | Linear | Linear | Linear |
| Warmup ratio | 0.06 | 0.06 | 0.06 | 0.05 | 0.05 |
| Num epochs | 20 | 15 | 10 | 20 | 20 |

Table 7: Fine-tuning hyperparameters for the RoBERTa language model on different tasks. TC: Text Classification, SA: Sentiment Analysis, QA: Question Answering, NER: Named Entity Recognition, SS: Semantic Similarity.

| Hyperparameter | TC | SA | QA | NER | SS |
|---|---|---|---|---|---|
| Max sequence length | 128 | 128 | 256 | 128 | 128 |
| Batch size | 8 | 8 | 2 | 8 | 8 |
| Learning rate | 2e-5 | 2e-5 | 2.e-5 | 1e-5 | 1e-5 |
| Num steps | 6K | 6K | 4K | 6K | 6K |
| Learning rate decay | Linear | Linear | Linear | Linear | Linear |

Table 8: Fine-tuning hyperparameters for the XLNet language model on different tasks. TC: Text Classification, SA: Sentiment Analysis, QA: Question Answering, NER: Named Entity Recognition, SS: Semantic Similarity.

| Hyperparameter | TC | SA | QA | NER | SS |
|---|---|---|---|---|---|
| n_highway | 2 | 2 | 2 | 2 | 2 |
| Droupout | 0.2 | 0.2 | 0.2 | 0.2 | 0.2 |
| Batch size | 128 | 128 | 256 | 128 | 128 |
| Projection dim | 512 | 512 | 512 | 512 | 512 |
| Num epochs | 20 | 20 | 10 | 20 | 20 |

Table 9: Fine-tuning hyperparameters for the ELMo language model on different tasks. TC: Text Classification, SA: Sentiment Analysis, QA: Question Answering, NER: Named Entity Recognition, SS: Semantic Similarity.

# Appendix B. Manual curation of perturbed samples

Table 10 shows how many perturbed samples (*word-level deletion*) were checked in the manual curation procedure until reaching 200 test samples for every dataset and different values of Perturbation Per Sample (PPS). Table 11 and Table 12 show the same statistics for *word-level negation* and *word-level replacement with synonym*, respectively.

| Dataset | Perturbation Per Samples | | | |
|---|---|---|---|---|
| | PPS=1 | PPS=2 | PPS=3 | PPS=4 |
| TREC | 221 | 269 | 341 | 435 |
| SST | 238 | 301 | 345 | 452 |
| CoNLL | 261 | 327 | 394 | 468 |
| STS | 240 | 281 | 354 | 409 |
| WQA | 219 | 248 | 283 | 351 |

Table 10: The number of perturbed samples (word-level deletion) checked in the manual curation procedure until reaching 200 test samples for every dataset and different values of Perturbation Per Sample (PPS).

| Dataset | Perturbation Per Samples | | | |
|---|---|---|---|---|
| | PPS=1 | PPS=2 | PPS=3 | PPS=4 |
| TREC | 235 | 251 | 268 | 268 |
| SST | 291 | 317 | 325 | 325 |
| CoNLL | 209 | 218 | 226 | 226 |
| STS | 253 | 269 | 291 | 291 |
| WQA | 295 | 314 | 317 | 317 |

Table 11: The number of perturbed samples (word-level negation) checked in the manual curation procedure until reaching 200 test samples for every dataset and different values of Perturbation Per Sample (PPS).

| Dataset | Perturbation Per Samples | | | |
|---|---|---|---|---|
| | PPS=1 | PPS=2 | PPS=3 | PPS=4 |
| TREC | 239 | 266 | 295 | 308 |
| SST | 221 | 249 | 273 | 290 |
| CoNLL | 213 | 228 | 236 | 251 |
| STS | 230 | 251 | 284 | 303 |
| WQA | 215 | 233 | 256 | 287 |

Table 12: The number of perturbed samples (word-level replacement with synonym) checked in the manual curation procedure until reaching 200 test samples for every dataset and different values of Perturbation Per Sample (PPS).